\documentclass[runningheads]{llncs}
\usepackage[T1]{fontenc}
\usepackage{graphicx}
\usepackage{units}
\usepackage{tikz}
\usetikzlibrary{automata, positioning, arrows}
\tikzset{
->, 
node distance=3cm, 
every state/.style={thick, fill=gray!10, minimum size=1.5cm}, 
initial text=$ $, 
}

\usepackage{amsmath}

\begin{document}
%
%
\title{Controlling Robot Swarm Aggregation through a Minority of Informed Robots}
%
%
\titlerunning{Controlling Robot Swarm Aggregation with Informed Robots}
%
%
\author{Antoine Sion\inst{1}\orcidID{0000-0002-9901-3906} \and
Andreagiovanni Reina\inst{2}\orcidID{0000-0003-4745-992X} \and
Mauro Birattari\inst{2}\orcidID{0000-0003-3309-2194} \and
Elio Tuci\inst{1}\orcidID{0000-0001-7345-671X}}
%
\authorrunning{A. Sion et al.}
%
\institute{Faculty of Computer Science, University of Namur, Belgium \email{\{antoine.sion,elio.tuci\}@unamur.be} \and IRIDIA, Université
Libre de Bruxelles, Brussels, Belgium
\email{\{andreagiovanni.reina,mauro.birattari\}@ulb.be}}
%
\index{Sion, Antoine}
\index{Reina, Andreagiovanni}
\index{Birattari, Mauro}
\index{Tuci, Elio}
%
\maketitle              
\begin{abstract}

Self-organized aggregation is a well studied behavior in swarm robotics as it is the pre-condition for the development of more advanced group-level responses. In this paper, we investigate the design of decentralized algorithms for a swarm of heterogeneous robots that self-aggregate over distinct target sites. A previous study has shown that including as part of the swarm a number of informed robots can steer the dynamic of the aggregation process to a desirable distribution of the swarm between the available aggregation sites. We have replicated the results of the previous study using a simplified approach: we removed constraints related to the communication protocol of the robots and simplified the control mechanisms regulating the transitions between states of the probabilistic controller. The results show that the performances obtained with the previous, more complex, controller can be replicated with our simplified approach which offers clear advantages in terms of portability to the physical robots and in terms of flexibility. That is, our simplified approach can generate self-organized aggregation responses in a larger set of operating conditions than what can be achieved with the complex controller.

\end{abstract}

\section{Introduction}
\label{sec:intro}

Swarm robotics can be defined as ``the study of how a large number of relatively simple physically embodied agents can be designed such that a desired collective behavior emerges from the local interactions among the agents and between the agents and the environment''~\cite{Sahin2008}. Robot swarms aim to be robust, flexible and scalable due to their decentralized nature and their large group size. Brambilla et al.\,\cite{Brambilla2013SwarmRoboticsAReviewFromTheSwarmEngineeringPerspective} and Schranz et al.\,\cite{Schranz2020SwarmRoboticBehaviorsAndCurrentApplications} give an overview of the basic collective behaviors that robot swarms can display in order to achieve complex tasks such as coordinated motion, task allocation, self-assembly or aggregation, which is studied in this paper. Aggregation is used to group the swarm in a location---for example, to initiate a collaborative task that requires physical proximity among the robots. Inspiration for self-organized aggregation can be found in animals such as bees~\cite{Szopek2013} or cockroaches~\cite{Jeanson2005}.

Several approaches have been studied in swarm robotics to achieve aggregation. A popular method is to control the robots through probabilistic finite-state machines. Soysal and \c{S}ahin \cite{Soysal2005} analyzed the performance of a three-state finite-state machine while varying different transitional probabilistic parameters. Variations of the controller yielded different aggregation behaviors highlighting the importance of the evaluation metrics used such as cluster sizes or total distances between robots. Cambier et al.\,\cite{Cambier2021CulturalEvolutionOfProbabilisticAggregation} studied the influence of cultural propagation in an aggregation scenario. Using a simple finite-state machine with two states, robots communicated with each other to modify the probabilistic parameters of their controllers. This behavioral plasticity has ensured adaptability to dynamic environments and performed better than classical approaches. Simpler methods have also been proposed to achieve self-organized aggregation as in~\cite{Gauci2014SelfOrganizedAggregationWithoutComputation} where the only sensor available to the robots is a binary sensor informing them on the presence of another robot in their line of sight. The controller did not require computation and the parameters were chosen via a grid search. This approach is validated with successful experiments with 1000 simulated robots and 40 physical robots.  Francesca et al.\,\cite{Francesca2014AutoMoDeANovelApproachToTheAutomaticDesign} showed that self-organized aggregation can be achieved by robots controlled by probabilistic finite-state machines that are generated automatically. 

Artificial evolution has also been used to evolve controllers resulting in self-organized aggregation~\cite{Dorigo2004}.
%
Francesca et al.\,\cite{Francesca2012AnalysingAnEvolvedRoboticBehaviourUsingABiologicalModelOfCollegialDecisionMaking} evolved a neural network controller to perform self-organized aggregation and showed that the dynamics of the resulting collective behavior closely matches the one of a biological model that describes how cockroaches select a resting shelter.
Kengyel et al.~\cite{Kengyel2015PotentialOfHeterogeneityInCollectiveBehaviors} studied aggregation in heterogeneous robot swarms where each robot ran one among four distinct behaviors inspired from aggregation in honeybees. The number of robots running the alternative behaviors were varied using an evolutionary algorithm. The results of the simulations show that through the cooperation of different types of behaviors, the heterogeneous swarm performs better than its homogeneous equivalent. 

In this study, we revisit the methods illustrated in~\cite{Firat2020Group-sizeRegulationInSelf-OrganisedAggregation} to steer self-organized aggregation responses with the use of informed robots. Generally speaking, the use of informed individuals to steer the dynamic process leading to the group response is a method to guide self-organisation in distributed systems inspired by behaviors observed in biology. In a seminal study, Couzin et al.\,\cite{CouzinEtAl2005} studied the influence of informed individuals in collective decision making in the context of collective animal motion. The informed individuals had a preferred direction of motion and biased the collective decision in that direction. The rest of the swarm did not have any preferred direction of motion, nor was able to recognise informed individuals as such. The study shows that the accuracy of the group motion towards the direction known by the informed agents increases asymptotically as the proportion of informed individuals increases, and that the larger the group, the smaller the proportion of informed individuals needed to guide the group with a given accuracy. A recent study on cockroaches~\cite{Martin2021ConsensusDrivenByAMinorityInHeterogeneousGroups} has shown that a minority of individuals preferring one shelter over the other can influence the population to reach a consensus for only one site. The technique of using informed individuals to steer the collective dynamics has already been used in artificial systems with various types of collective behaviors. For example, informed individuals have been used in flocking~\cite{Celikkanat-Sahin2010SteeringSelf-organizedRobotFlocksThroughExternallyGuidedIndividuals,Ferrante2012Self-organizedFlockingWithAMobileRobotSwarmANovelMotionControlMethod,Ferrante2014ASelf-adaptiveCommunicationStrategyForFlockingInStationaryAndNonStationaryEnvironments} to guide the robot swarm in the desired direction, in collective decision making~\cite{Prasetyo2019CollectiveDecisionMakingInDynamicEnvironments,Masi2021} to achieve adaptability, and in self-organized aggregation to differentiate between multiple sites~\cite{Firat2020OnSelf-OrganisedAggregationDynamics,Firat2020Group-sizeRegulationInSelf-OrganisedAggregation,Firat2018TPNC,Gillet2019GuidingAggregationDynamicsInASwarmOfAgentsViaInformedIndividuals}.


In this paper, we replicate the experimental setup originally illustrated in~\cite{Firat2020Group-sizeRegulationInSelf-OrganisedAggregation} by showing that equally effective aggregation dynamics can be obtained with a largely simplified approach. The aggregation response described in~\cite{Firat2020Group-sizeRegulationInSelf-OrganisedAggregation} took place in a circular arena with two aggregation sites, the black and the white site (see Fig.~\ref{fig:methods}a). The goal of the swarm was to distribute the robots between the two sites in a desired proportion (e.g. 70\% of the robots on the white site and the remaining 30\% on the black one) while minimizing the total proportion of informed robots. Non-informed robots treated both sites in the same way while informed robots were designed to systematically avoid one of the sites and to rest on the other. The swarm comprised a certain proportion of informed robots, among which there were those that selectively rested on the black site and those that rested only on the white site. The study shows that the proportion of robots on black/white site at the end of an aggregation process matches the proportion of informed robots resting only on the black/white site relative to the number of informed robots in the swarm (see Sec.~\ref{sec:methods} for more details). These aggregation dynamics can be achieved using a relatively low percentage of informed robots in the swarm---roughly 30\% for a swarm of 100 robots. 

In~\cite{Firat2020Group-sizeRegulationInSelf-OrganisedAggregation}, the robots were controlled by a probabilistic finite-state machine illustrated in details in Sec.~\ref{sec:methods} and depicted in Fig.~\ref{fig:methods}c. In this study, we question the nature of the mechanisms regulating the transition from state Stay ($\mathcal{S}$) to state Leave ($\mathcal{L}$), and the type of communication protocol involved in these mechanisms, by suggesting an alternative solution that largely simplifies the rule regulating the transition between these two states for non-informed robots. Following the ``Occam's razor'' principle, the solution we illustrate in this paper can represent a preferable alternative to the more complex solution proposed in~\cite{Firat2020Group-sizeRegulationInSelf-OrganisedAggregation} and can help the porting of the control system to the physical robots. Moreover, we remove differences both in behavior and in communication capabilities between informed and non-informed robots that, in our view, undermine the robustness of the aggregation process. For example, in~\cite{Firat2020Group-sizeRegulationInSelf-OrganisedAggregation}, non-informed robots rest on an aggregation site if this is already inhabited by at least one informed robot that signals its presence on the site. Thus, an aggregation process can only be initiated by informed robots. This can significantly delay the aggregation process, in particular in those operating conditions in which the probability that an informed robot finds and rests on an aggregation site is relatively low. Additionally, this makes the swarm less robust to the loss of informed robots. In the extreme case of a loss of all informed robots, there will be no aggregation at all. With our approach, aggregation dynamics can emerge independently of the presence of informed robots. This increases the range of group responses that the swarm can generate without interfering with the effectiveness of the aggregation dynamics. 

\section{Materials and Methods}
\label{sec:methods}
\begin{figure}[t]
\centering
\begin{tabular}{ccc}
\includegraphics[width=0.3\textwidth]{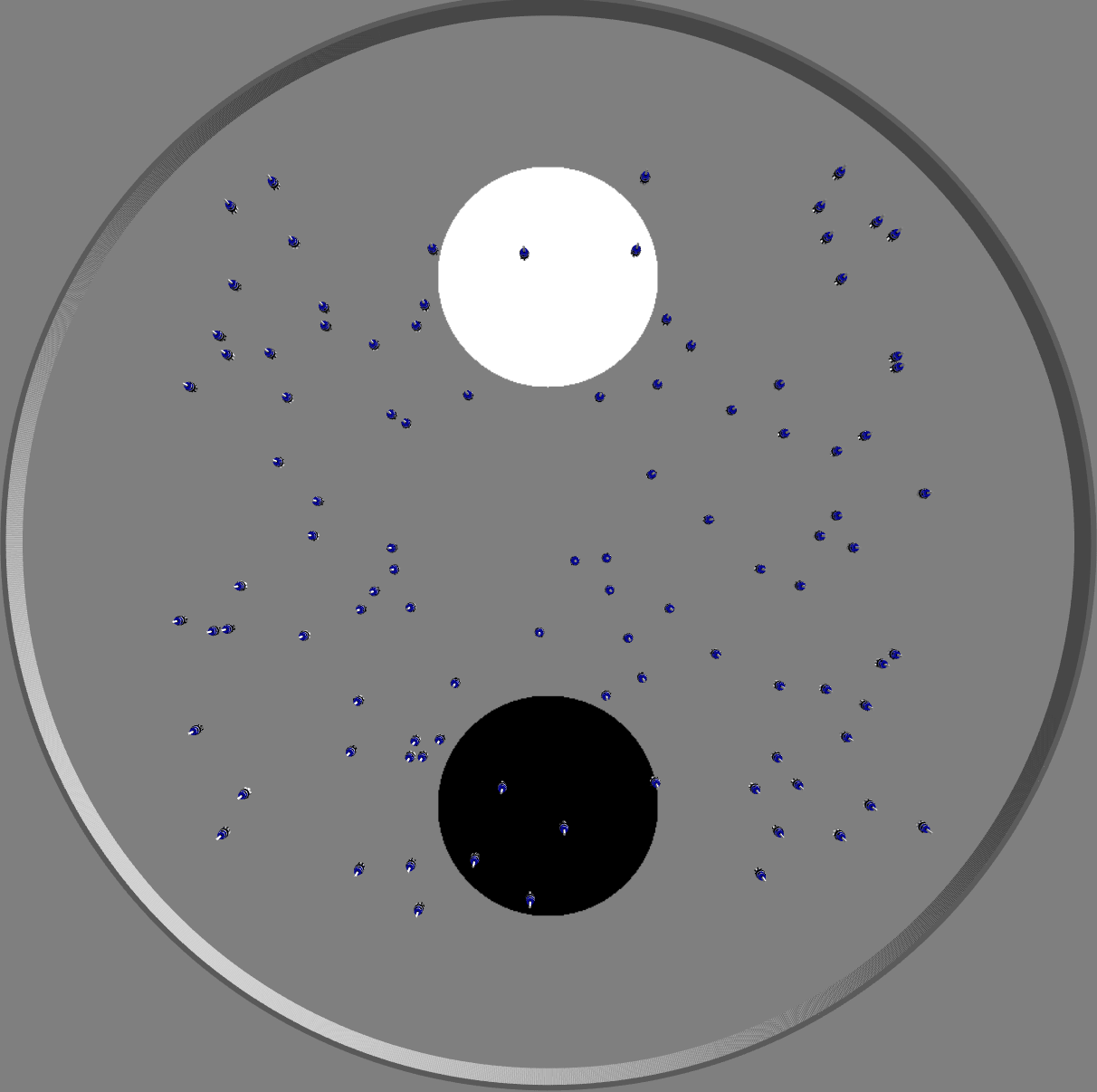}&~~
\includegraphics[width=0.165\textwidth]{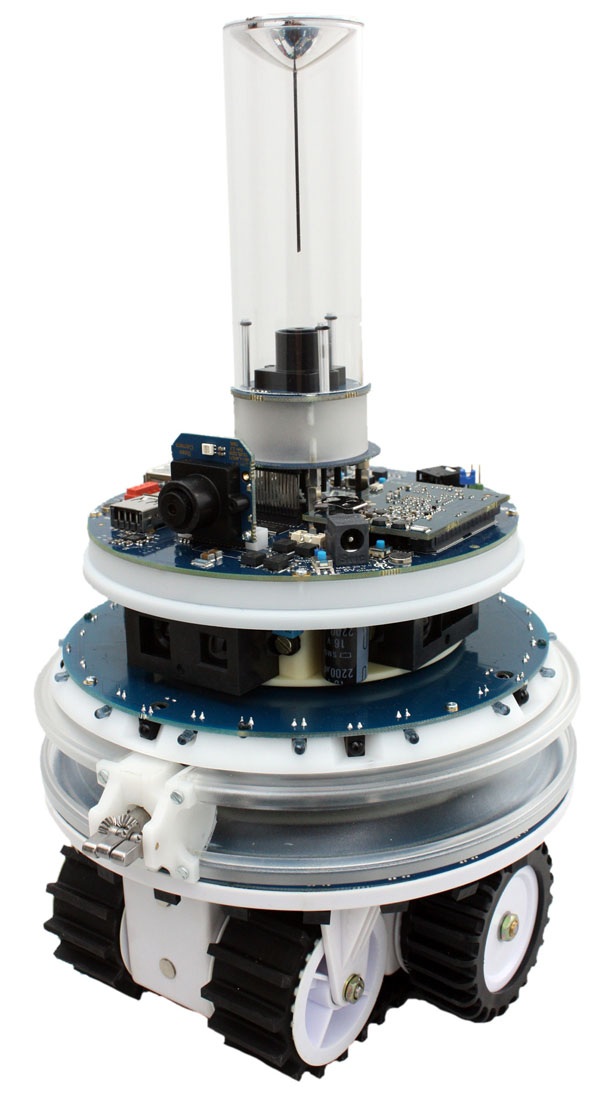}&~~
\includegraphics[width=0.35\textwidth]{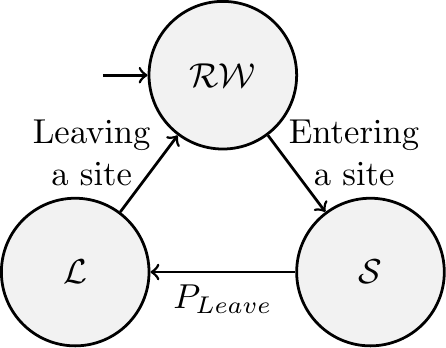}\\
(a) & (b) & (c)
\end{tabular}
\caption{(a) The simulated circular arena with the robots (blue dots) and the aggregation sites (black and white circles). (b) The Foot-bot robot \cite{bonani2010marxbot}; image from \url{www.swarmanoid.org}. (c) The probabilistic finite-state machine controlling the robot behavior composed of three states; $\mathcal{RW}$ (Random Walk), $\mathcal{S}$ (Stay) and $\mathcal{L}$ (Leave).}
\label{fig:methods}
\end{figure}
The experiments are ran using ARGoS~\cite{Pinciroli2012ARGoSAModularParallelMulti}, which is one of the best-performing simulators to run physics-based simulations for large-scale robot swarms~\cite{Pitonakova2018}. The robots operate in a circular environment where the aggregation sites are circular areas located at equal distance between the center of the arena and its perimetric wall (Fig.~\ref{fig:methods}a). One site is painted in black, the other in white; the arena floor is grey. We perform our experiments with a simplified model of the Foot-bot \cite{bonani2010marxbot} (see Fig.~\ref{fig:methods}b), a mobile robot that we equip with three types sensors: a ground sensor, an array of proximity sensors, and the range-and-bearing sensor. The ground sensor is located on the robot's underside, to detect the color of the ground; it returns 1, 0.5, and 0 when a robot is on a white, grey, and black floor, respectively. The proximity sensors use infrared (IR) signals to detect obstacles such as other robots or the arena walls. The range-and-bearing sensor is used by the robots to exchange simple communication signals when they are located at less than \unit[0.8]{m} from each other.

A trial starts with the robots randomly placed in the arena following a uniform distribution, and it terminates after \unit[30,000]{s}. As in~\cite{Firat2020Group-sizeRegulationInSelf-OrganisedAggregation}, the swarm of size $N$ is composed of informed robots (in proportion $\rho_{I}$) and non-informed ones (in proportion $ 1-\rho_{I}$). During a trial, both informed and non-informed robots move randomly in the environment at \unitfrac[10]{cm}{s}. Non-informed robots can rest on any aggregation site as soon as they enter into one of them. Informed robots selectively avoid to rest on one site and rest only on the other one. There are informed robots for black---which avoid to rest on the white site, and rest on the black aggregation site only---and informed robots for white---which avoid to rest on the black site, and rest on the white site only. Let $N_{sb}$ be the number of informed robots for black, $N_{sw}$ the number of informed robots for white, $\rho_{sb} = \frac{N_{sb}}{N_{sb}+N_{sw}}$ the proportion of informed robots for black relative to the total number of informed robots, and $\rho_{sw} = 1 - \rho_{sb}$ the proportion of informed robots for white relative to the total number of informed robots. 
Firat et al.\,~\cite{Firat2020Group-sizeRegulationInSelf-OrganisedAggregation} demonstrated that, with a relatively small proportion of informed robots in the swarm ($\rho_{I} \approx 0.3$) and for different values of $\rho_{sb}$ and $\rho_{sw}$, the number of robots aggregated on the black site is approximately equal to $N \, \rho_{sb} $, while the number of those aggregated on the white site is approximately equal to $N \, \rho_{sw}$, for swarms of size $N=50$ and for $N=100$. Our objective is to replicate this aggregation dynamics with a largely simplified model illustrated below.

The primary difference between our model and the one introduced in~\cite{Firat2020Group-sizeRegulationInSelf-OrganisedAggregation} resides in the way in which the robots communicate while within an aggregation site. In~\cite{Firat2020Group-sizeRegulationInSelf-OrganisedAggregation}, communication is needed by the robots to count how many informed robots are resting on the aggregation site within the communication range. In our model, communication is needed by the robots to count how many robots (including both informed and non-informed) are resting on the aggregation site within the communication range. This small difference between the two models derives from a rather substantial modification of the robots' communication system which we apply to the original model as illustrated in~\cite{Firat2020Group-sizeRegulationInSelf-OrganisedAggregation} to improve the robustness and flexibility of the swarm's behavior. In~\cite{Firat2020Group-sizeRegulationInSelf-OrganisedAggregation}, informed robots emits signals (i.e., one bit signal), while non-informed robots can only receive these signals, they can not emit them. Contrary to~\cite{Firat2020Group-sizeRegulationInSelf-OrganisedAggregation}, in our model, informed and non-informed robots are functionally identical with respect to communication; they can both send and receive signals. This implies that, contrary to~\cite{Firat2020Group-sizeRegulationInSelf-OrganisedAggregation}, in our model communication signals indicate only the presence of spatially proximal robots within a site without saying anything about the identity of the signal's sender (i.e., whether it is an informed or a non-informed robot). Therefore, the results of our study can generalise to application scenarios where the communication is indirect, i.e., the robots count the neighbors in its view range without the need of an exchange of messages and without the need of distinguishing between informed and non-informed robots.

Both in~\cite{Firat2020Group-sizeRegulationInSelf-OrganisedAggregation} and in our model, the robots are controlled by a probabilistic finite state-machine (hereafter PFSM, see Fig.~\ref{fig:methods}c), which is updated every 2 seconds and comprises three states: Random Walk ($\mathcal{RW}$), Stay ($\mathcal{S}$), and Leave ($\mathcal{L}$). However, in our model, the rules that regulate the transition between different states have been modified with respect to the original implementation as illustrated in~\cite{Firat2020Group-sizeRegulationInSelf-OrganisedAggregation}. In the following, we describe the PFSM and we illustrate the modifications we introduced with respect to~\cite{Firat2020Group-sizeRegulationInSelf-OrganisedAggregation}.

At the beginning of a trial, the robots are in state $\mathcal{RW}$. They explore the environment while avoiding obstacles with an isotropic random walk based on straight motion and random rotation. The robots move straight for \unit[5]{s} at a speed of \unitfrac[10]{cm}{s}, and turn with turning angles taken from a wrapped Cauchy distribution \cite{AnExtendedFamilyOfCircularDistributionsRelatedToWrappedCauchy}. The probability density function is the following :
\begin{equation}
    f_{\omega}(\theta, \mu, \rho)=\frac{1}{2 \pi} \frac{1-\rho^{2}}{1+\rho^{2}-2 \rho \cos (\theta-\mu)}, \quad 0<\rho<1,
\end{equation}
where $\mu$ is the average value of the distribution and $\rho$ the skewness. With $\rho = 0$, the wrapped Cauchy distribution becomes uniform and there is no correlation between the movement directions before and after a turn. With $\rho = 1$, we have a Dirac distribution and the robot follows a straight line. Here we take $\rho =0.5$. During this behavior, when the proximity sensors detect an obstacle (the wall or other robots), the robot stops and turns of an angle chosen uniformly in the interval $[-\pi,\pi]$. After turning, if there is no obstruction ahead, the robot resumes its normal random walk otherwise, it repeats the manoeuvre.

Both in our model and in~\cite{Firat2020Group-sizeRegulationInSelf-OrganisedAggregation}, informed robots systematically transition from state $\mathcal{RW}$ to the state $\mathcal{S}$ when they enter their preferred site, otherwise they move randomly. While in $\mathcal{S}$, informed robots rest on the aggregation site. The condition that triggers the transition from state $\mathcal{RW}$ to the state $\mathcal{S}$ for non-informed robots is different in our model with respect to~\cite{Firat2020Group-sizeRegulationInSelf-OrganisedAggregation}. In particular, in~\cite{Firat2020Group-sizeRegulationInSelf-OrganisedAggregation},  non-informed robots switch to state $\mathcal{S}$ if, while entering into an aggregation site, they perceive the presence of informed robots at the site. Contrary to~\cite{Firat2020Group-sizeRegulationInSelf-OrganisedAggregation}, in our approach, non-informed robots systematically transition from state $\mathcal{RW}$ to state $\mathcal{S}$ whenever they enter into an aggregation site regardless of the presence of any other type of robot at the site. In both studies, when a robot enters an aggregation site, it continues moving forward for 10 seconds to avoid a congestion of robots on the perimeter of the site that could eventually hinder other robots from entering. Then, it rests on the site.

Both in our model and in~\cite{Firat2020Group-sizeRegulationInSelf-OrganisedAggregation}, informed robots never leave the state $\mathcal{S}$. That is, once an informed robot finds its preferred site, it never leaves the site. Non-informed robots transition from state $\mathcal{S}$ to state $\mathcal{L}$ with a probability that is computed differently with respect to~\cite{Firat2020Group-sizeRegulationInSelf-OrganisedAggregation}. In~\cite{Firat2020Group-sizeRegulationInSelf-OrganisedAggregation},
non-informed robots transition from state $\mathcal{S}$ to state $\mathcal{L}$ with a probability $P_{Leave}$ computed as
\begin{equation}
  P_{Leave} =
    \begin{cases}
      e^{-a(k-|n-x|)} & \text{if $n>0$,}\\
      1 & \text{if $n = 0$;}\\
    \end{cases}       
    \label{eq:oldPLeave}
\end{equation}
where $n$ and $x$ are the number of informed robots resting on the site within communication distance at this moment and at the moment of joining this site, respectively. Parameters $a$ and $k$ are fixed to $a = 2$ and $k = 18$. With Eq.~\eqref{eq:oldPLeave}, the transition from state $\mathcal{S}$ to state $\mathcal{L}$ is based on the temporal variation in the number of informed robots perceived at an aggregation site. 

Contrary to~\cite{Firat2020Group-sizeRegulationInSelf-OrganisedAggregation}, in our model the probability $P^{'}_{Leave}$ that regulates the transition from state $\mathcal{S}$ to state $\mathcal{L}$ for non-informed robots is computed as
\begin{equation}
P^{'}_{Leave} = \alpha e^{-\beta n} \;,
\label{secondPleave}
\end{equation}
with $n$ the number of robots (including both informed and non-informed) within communication distance, $\alpha = 0.5$, and $\beta = 2.25$\footnote[1]{The parameters $\alpha$ and $\beta$ have been fine-tuned to achieve a symmetry-breaking behavior in a homogeneous swarm of $N=100$ non-informed robots using the same arena setup illustrated in~\cite{Firat2020Group-sizeRegulationInSelf-OrganisedAggregation}.}. Contrary to~\cite{Firat2020Group-sizeRegulationInSelf-OrganisedAggregation}, $P^{'}_{Leave}$  relies on the fact that any type of robot in the swarm can broadcast and perceive communication signals while resting on a site. This transforms $P^{'}_{Leave}$ into something that depends on the estimated local density of (any type of) robots, while in~\cite{Firat2020Group-sizeRegulationInSelf-OrganisedAggregation} $P_{Leave}$ depends on the variation in the number of informed robots in the neighborhood. Hence, $P_{Leave}$ requires the robots to use memory and distinguish between types of robots, $P^{'}_{Leave}$ does not.

Both in~\cite{Firat2020Group-sizeRegulationInSelf-OrganisedAggregation} and in our approach a robot in state $\mathcal{L}$ exits an aggregation site by moving forward and avoiding obstacles. Once outside the site, it systematically transitions to state $\mathcal{RW}$. 

To summarise, the main differences between the model illustrated in~\cite{Firat2020Group-sizeRegulationInSelf-OrganisedAggregation} and our model are the following: (i)~in~\cite{Firat2020Group-sizeRegulationInSelf-OrganisedAggregation} informed robots only emit signals and non-informed robots only receive signals. In our models, both informed and non-informed robots send and receive communication signals. (ii)~In~\cite{Firat2020Group-sizeRegulationInSelf-OrganisedAggregation}, non-informed robots rest on a site only if they perceive the presence of informed robots, and leave a site with a probability that depends on the variation in the number of perceived informed robots. Hence, it requires memory of the past and a comparison between the present and the past state. In our model, non-informed robots systematically rest on an aggregation site, and they leave it with a probability that is determined by the current local density of robots (regardless of whether they are informed or non-informed) at the site. Therefore, our model is reactive and does not require any form of memory. In the next section, we show that the above mentioned modifications improve the robustness and the behavioral flexibility of the swarm. 

\section{Results}
\label{sec:results}
\begin{table}[b]
\caption{Parameters values}
\label{fig:parameters}
\begin{tabular}{|l|l|}\hline
Experiment parameters & Values \\\hline
Swarm size ($N$) & $\{ 50,100 \}$    \\ 
Proportion of informed robots ($\rho_{I}$) & $\{0.1,0.2,0.3,0.4,0.5\}$ \\ 
Proportion of black informed robots ($\rho_{sb}$) & $\{0.5,0.6,0.7,0.8,0.9,1\}$\\
Arena diameter & 12.9 m (for $N=50$), 19.2 m (for $N=100$)\\
Aggregation site diameter & 2.8 m (for $N=50$) , 4.0 m (for $N=100$)\\\hline
\end{tabular}
\end{table}
In this section, we compare the performance of swarms of robots controlled by the PFSM as originally illustrated in~\cite{Firat2020Group-sizeRegulationInSelf-OrganisedAggregation} and by our modified PFSM presented in Sec.~\ref{sec:methods}. We evaluate the two approaches in several different conditions given by all the possible combinations of the parameters' values listed in Table~\ref{fig:parameters}. In particular, we vary the swarm size ($N$), the proportion of informed robots in the swarm ($\rho_{I}$), and the proportion of black ($\rho_{sb}$) and white ($\rho_{sw}$) informed robots. As in~\cite{Firat2020Group-sizeRegulationInSelf-OrganisedAggregation}, while changing the swarm size, the swarm density has been kept constant by changing the diameter of the arena and of the aggregation sites (see Table~\ref{fig:parameters} for details).

We recall that the task of the robots is to distribute themselves on each site in a way that the robots resting on the black and white sites should be equal to $N \, \rho_{sb} $ and $N \, \rho_{sw}$, respectively, with the proportion of informed robots $\rho_{I}$ being as small as possible.
\begin{figure}[t]
  \centering
  \begin{tabular}{cccc} 
  \includegraphics[width=0.25\textwidth]{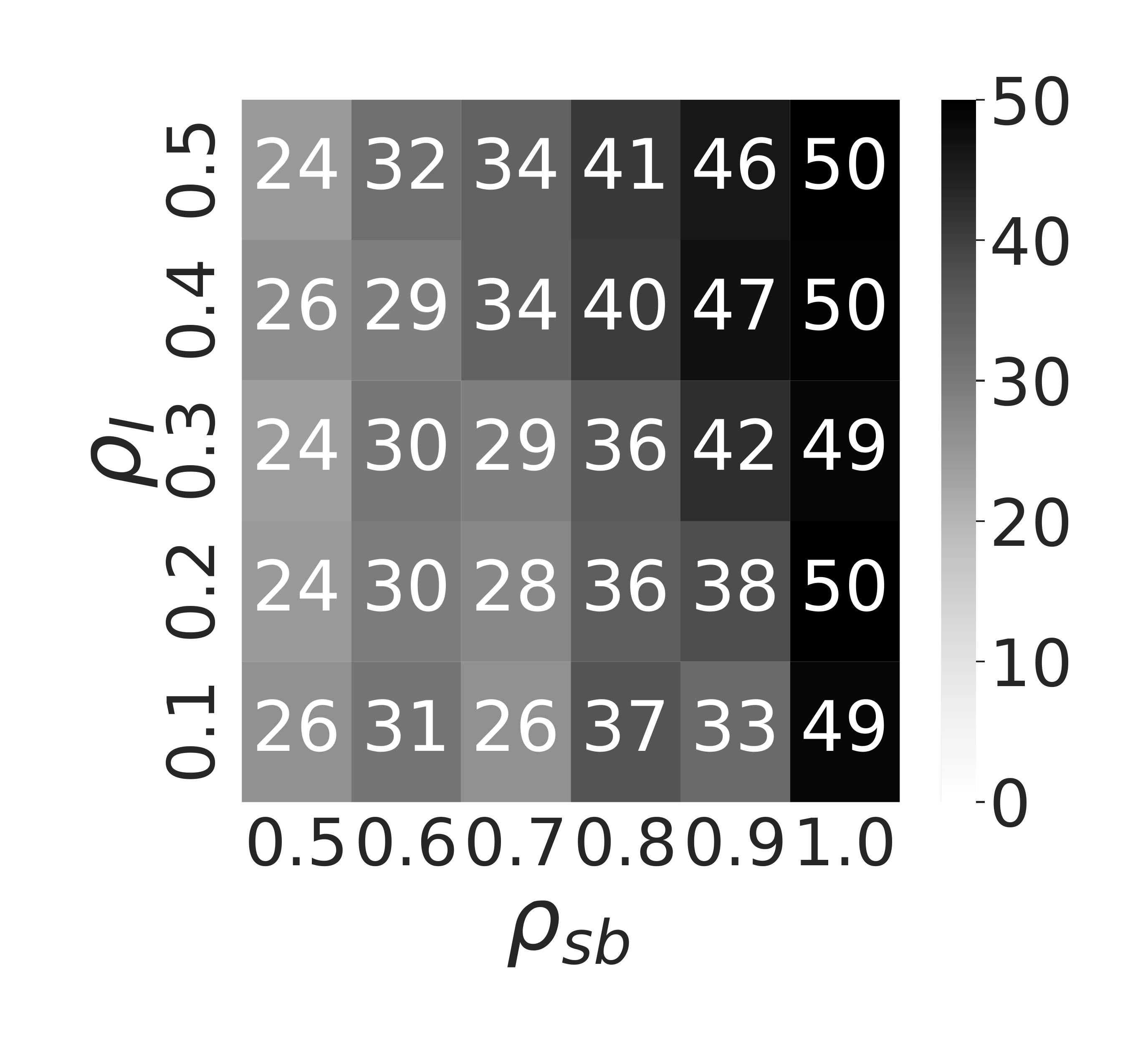}&
  \includegraphics[width=0.25\textwidth]{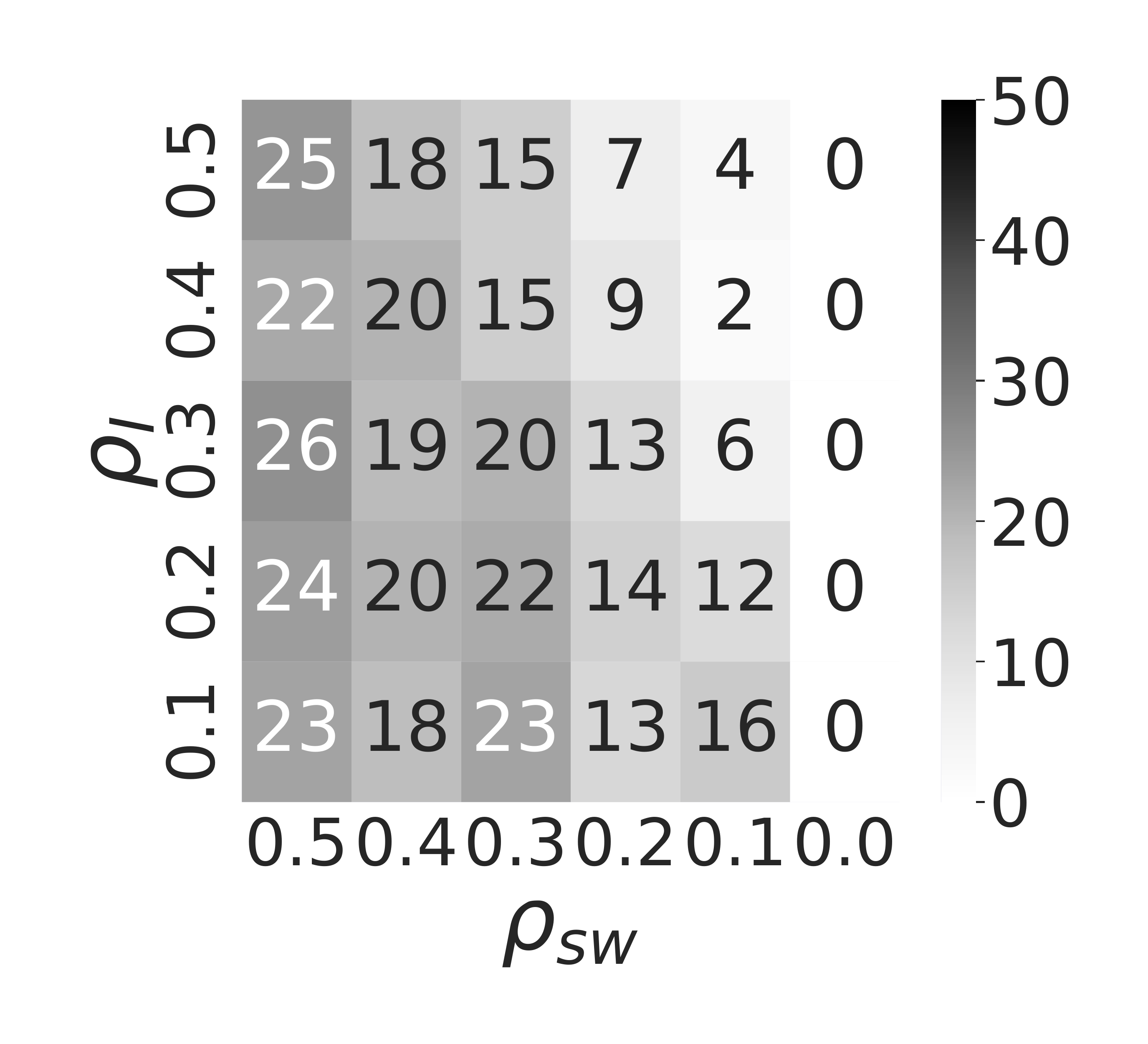}&
  \includegraphics[width=0.25\textwidth]{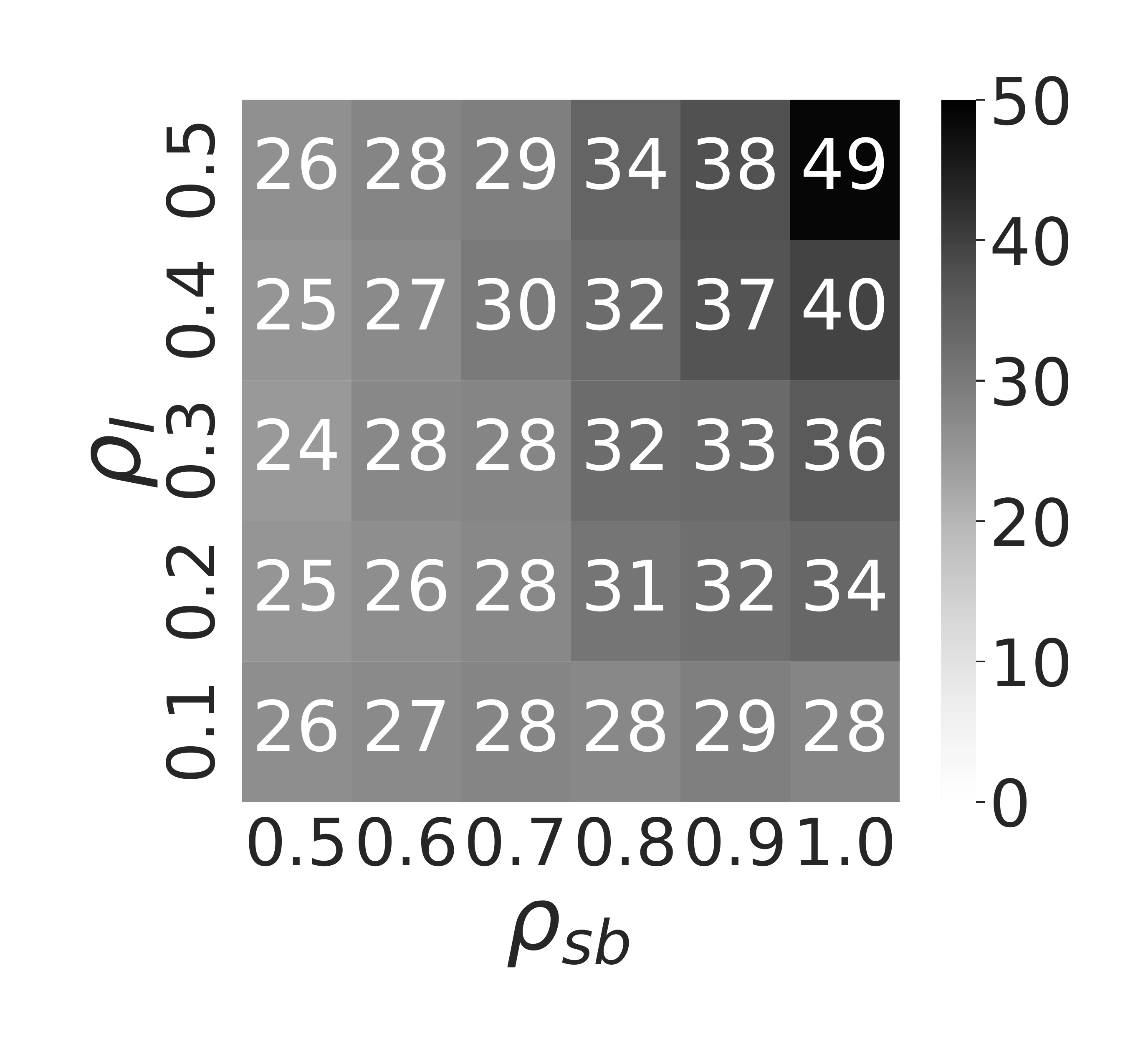}&
  \includegraphics[width=0.25\textwidth]{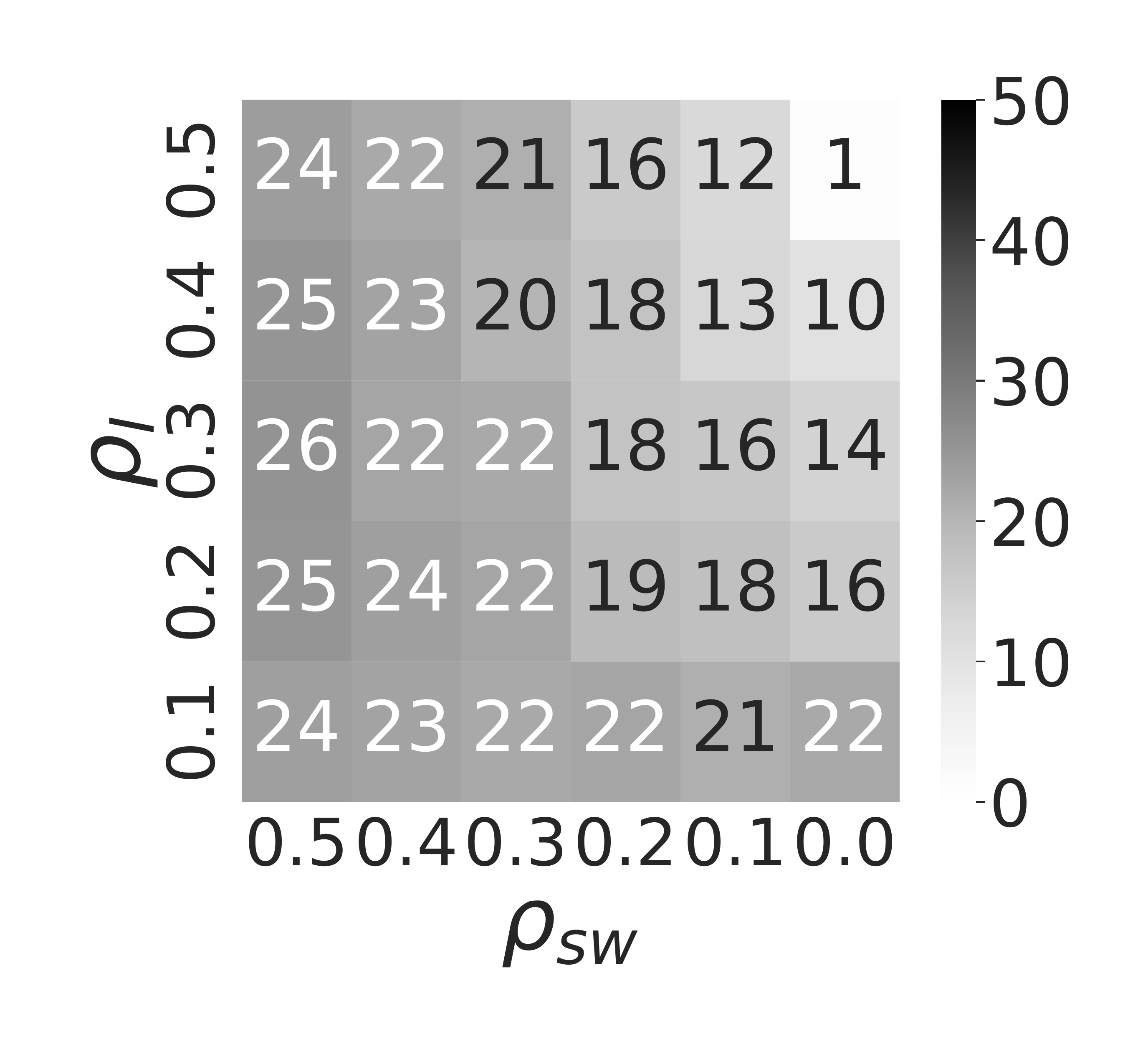}\\
  (a) & (b) & (c) & (d)\\
  \includegraphics[width=0.25\textwidth]{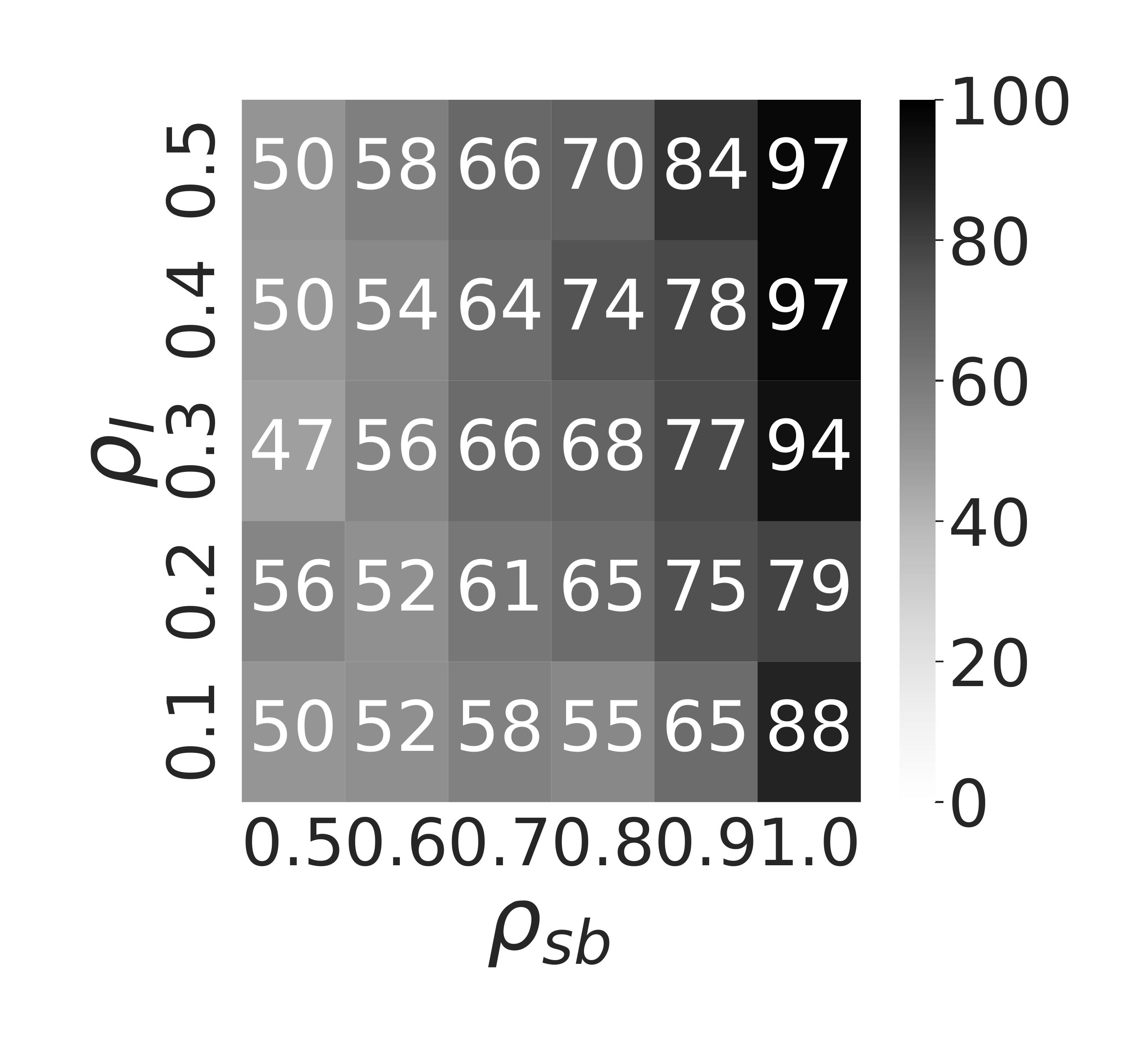}&
  \includegraphics[width=0.25\textwidth]{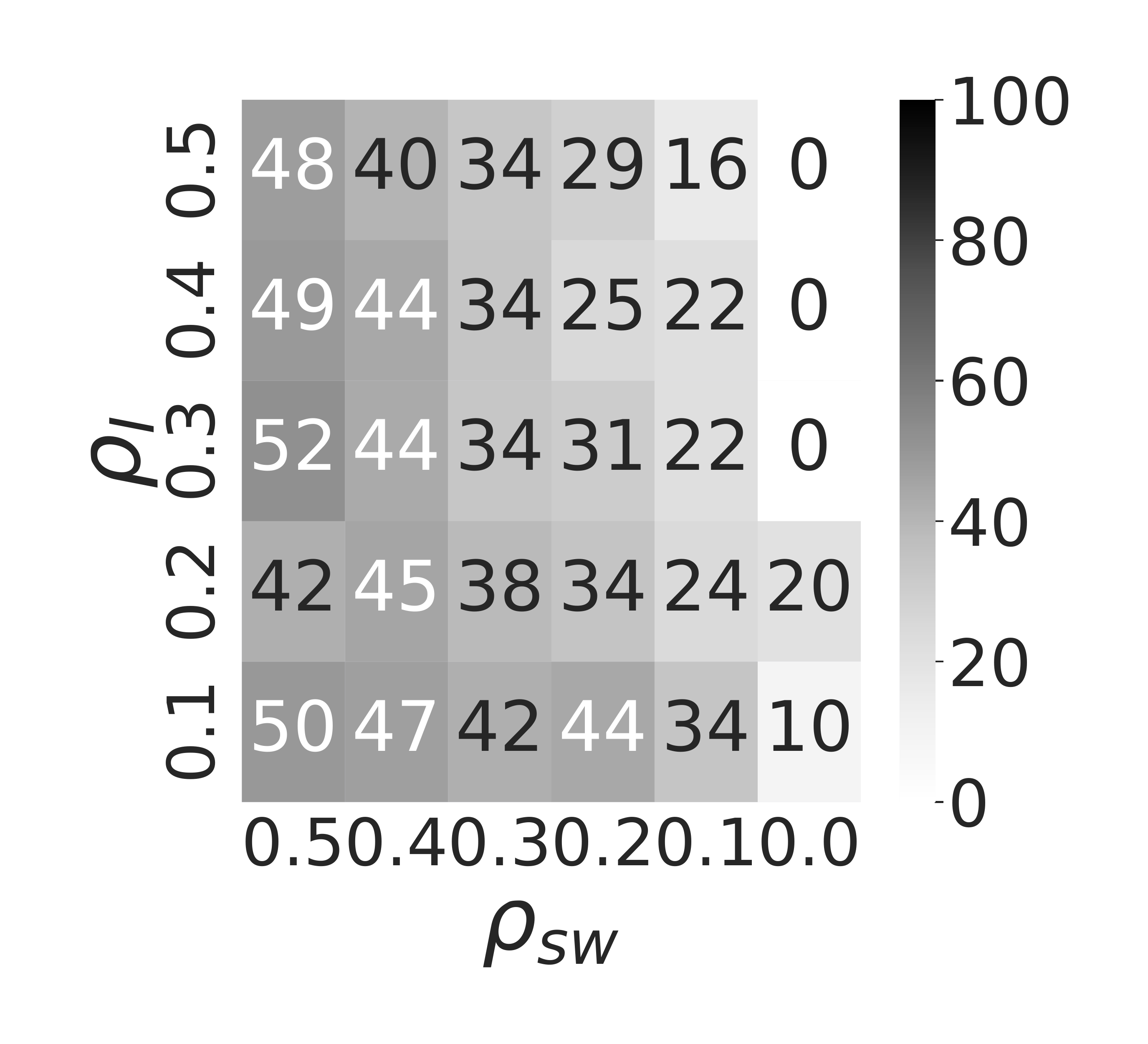}&
  \includegraphics[width=0.25\textwidth]{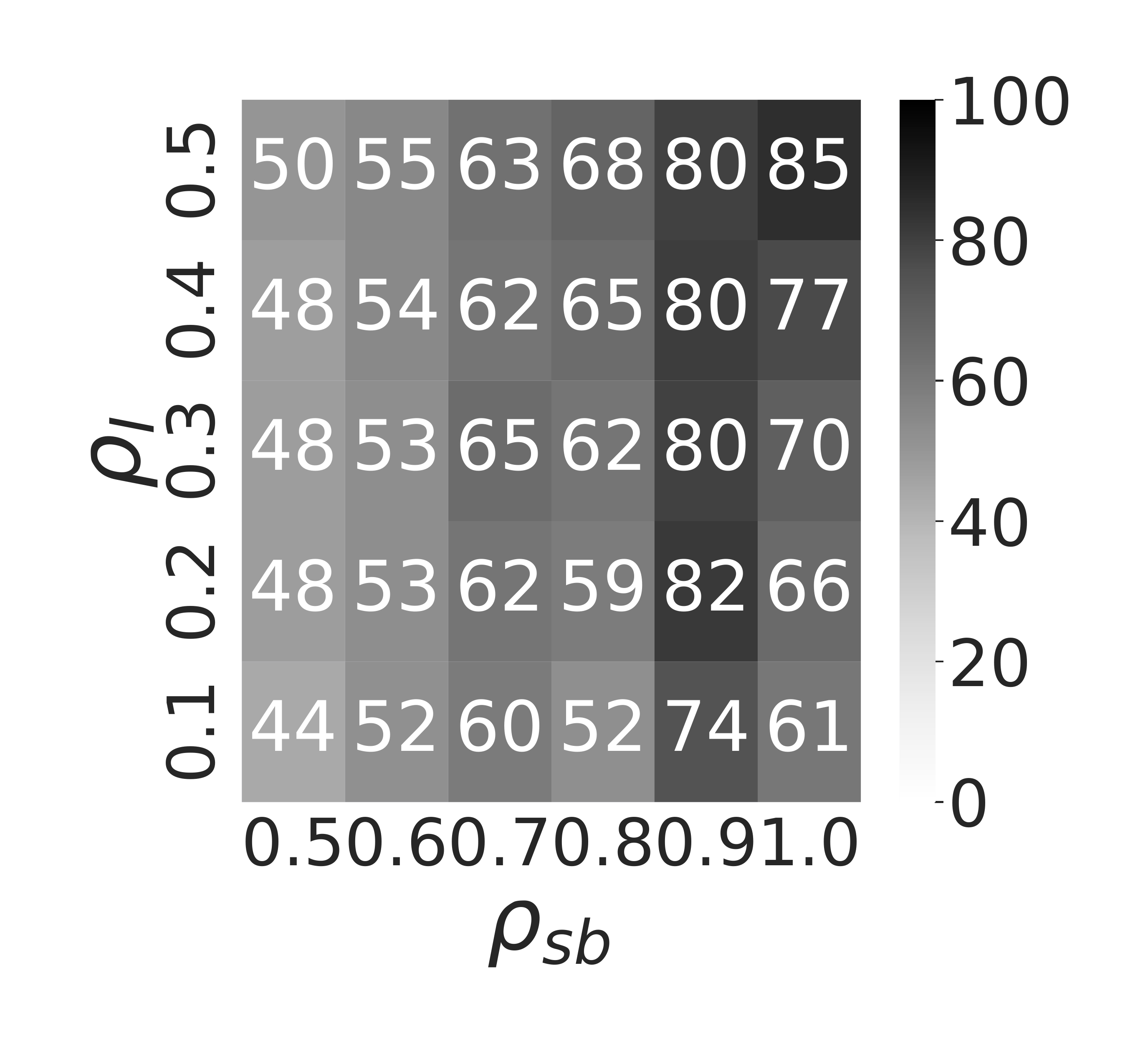}&
  \includegraphics[width=0.25\textwidth]{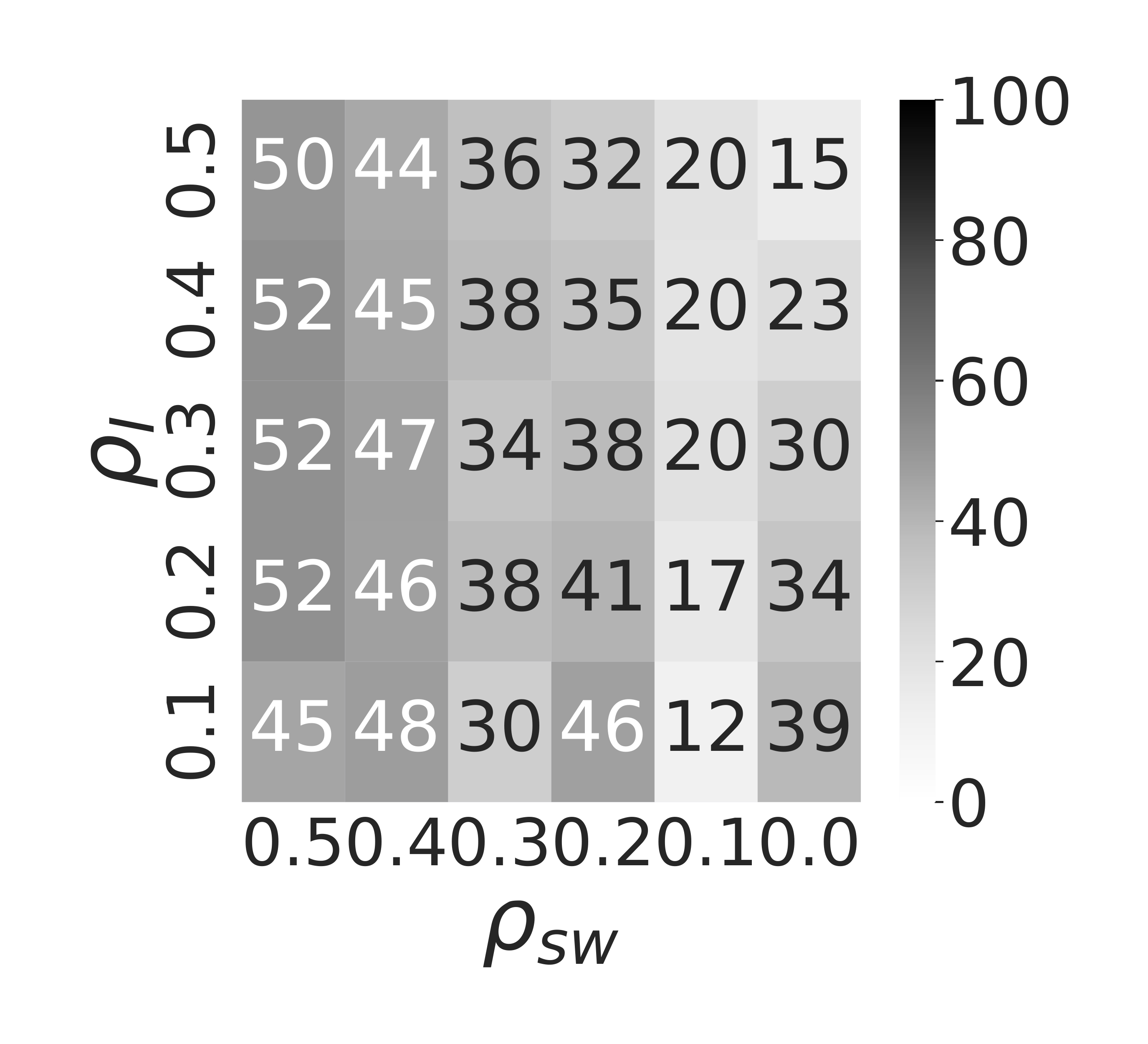}\\
    (e) & (f) & (g) & (h)
     \end{tabular}
     \caption{Graphs showing the median over 20 trials of the number of robots on each site at time \unit[$t = 30,000$]{s} (end of trials) for $N=50$ robots (top row) and $N=100$ robots (bottom row). For our approach the results are shown in (a) and (e) with reference to the black site, and in (b) and (f) with reference to the white site. For the approach in~\cite{Firat2020Group-sizeRegulationInSelf-OrganisedAggregation}, the results are shown in (c) and (g) with reference to the black site, and in (d) and (h) with reference to the white site. On all graphs, the y-axis refers to the proportion of informed robots in the swarm ($\rho_{I}$) and the shades of grey refer to the number of robots at each aggregation site with white indicating zero robots, and black indicating 50 robots in graphs on top row, and 100 robots in graphs on bottom row. The x-axis refers to ($\rho_{sb}$) in (a), (c), (e), and (g) and to ($\rho_{sw}$) in (b), (d), (f), and (h).}
     \label{fig:median}
\end{figure}
For each testing condition, and for each approach, we have run 20 trials. Figure~\ref{fig:median} shows, for both approaches, the median over 20 trials of the number of robots resting on each site at the end of each trial (i.e., at time \unit[$t=30,000$]{s}) for a swarm of $N=50$ robots (see Fig.~\ref{fig:median}, graphs in top row) and for a swarm of $N=100$ robots (see Fig.~\ref{fig:median}, graphs in bottom row). For our approach the results are shown in Figs.~\ref{fig:median}a and~\ref{fig:median}e with reference to the black site, and in Figs.~\ref{fig:median}b and~\ref{fig:median}f with reference to the white site. For the approach in~\cite{Firat2020Group-sizeRegulationInSelf-OrganisedAggregation}, the results are shown in Figs.~\ref{fig:median}c and~\ref{fig:median}g with reference to the black site, and in Figs.~\ref{fig:median}d and~\ref{fig:median}h with reference to the white site.

With a swarm of $N=50$ robots, both approaches perform relatively well, with the median of the number of robots resting on the black site increasing for progressively higher values of $\rho_{sb}$ (see Figs.~\ref{fig:median}a, and~\ref{fig:median}c) and decreasing for progressively lower values of $\rho_{sw}$ (see Figs.~\ref{fig:median}b and~\ref{fig:median}d). These trends can be clearly observed, for both approaches, for any values of $\rho_{I}$, even for the smallest tested value $\rho_{I}=0.1$ (i.e, 10\% of informed robots in the swarm). However, our approach performs better than the approach in~\cite{Firat2020Group-sizeRegulationInSelf-OrganisedAggregation} for values of $\rho_{I} > 0.3$ and $\rho_{sb} > 0.8$. In these conditions, our results are closer to the target robot distributions than the results obtained with the approach from~\cite{Firat2020Group-sizeRegulationInSelf-OrganisedAggregation}  (see Figs.~\ref{fig:median}a,~\ref{fig:median}b,~\ref{fig:median}c, and~\ref{fig:median}d, top right corners).
With a larger swarm size $N=100$, our results are closer to the target robot distributions for both $\rho_{I} < 0.3$ and $\rho_{sb} < 0.7$ (see Figs.~\ref{fig:median}e,~\ref{fig:median}f,~\ref{fig:median}g, and~\ref{fig:median}h, bottom left corners) and for $\rho_{I} > 0.3$ and $\rho_{sb} > 0.8$ (see Figs.~\ref{fig:median}e,~\ref{fig:median}f,~\ref{fig:median}g, and~\ref{fig:median}h, top right corners).

\begin{figure}[t]
  \centering
     \begin{tabular}{cccc}
         \includegraphics[width=0.25\textwidth]{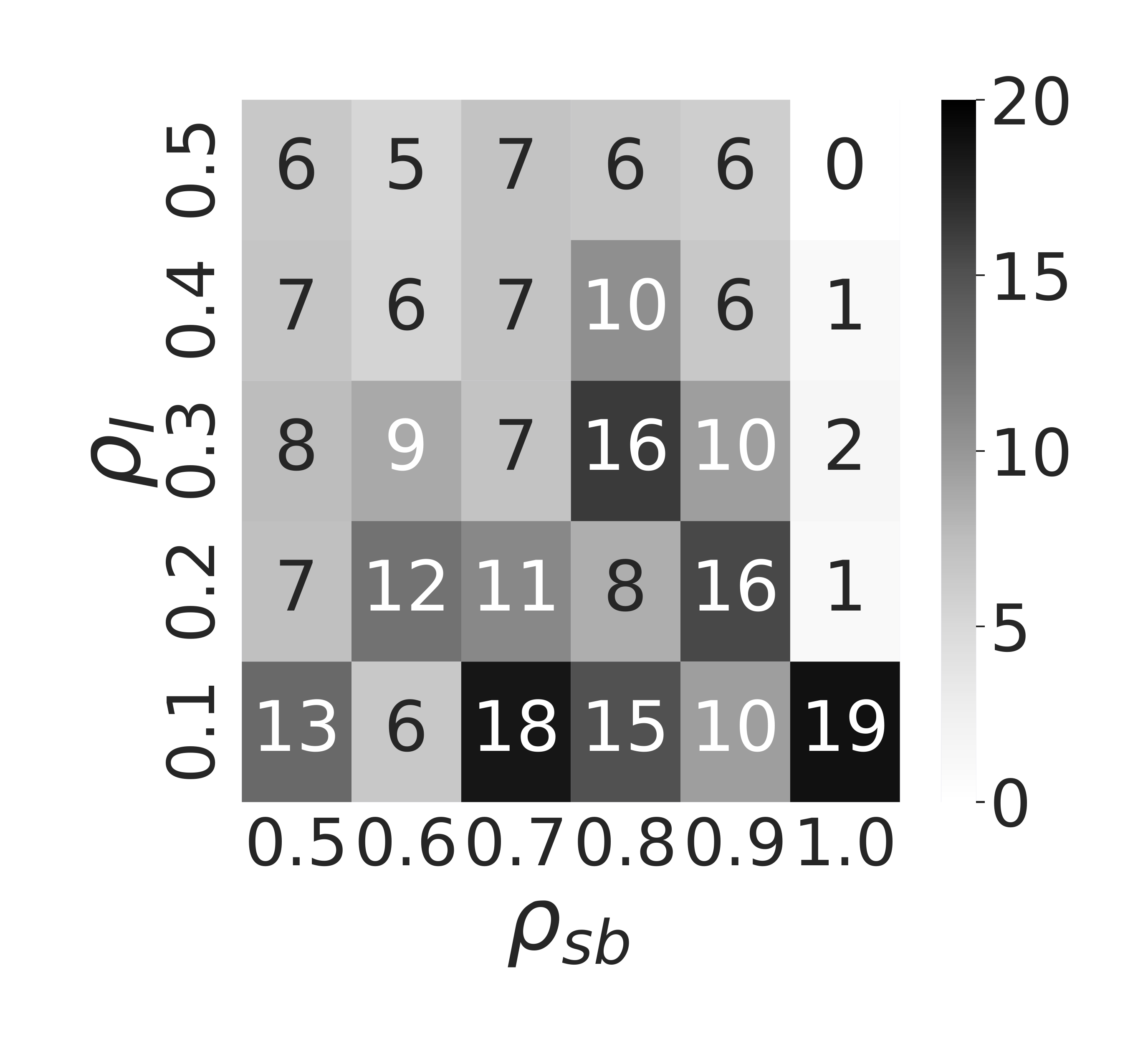}&
         \includegraphics[width=0.25\textwidth]{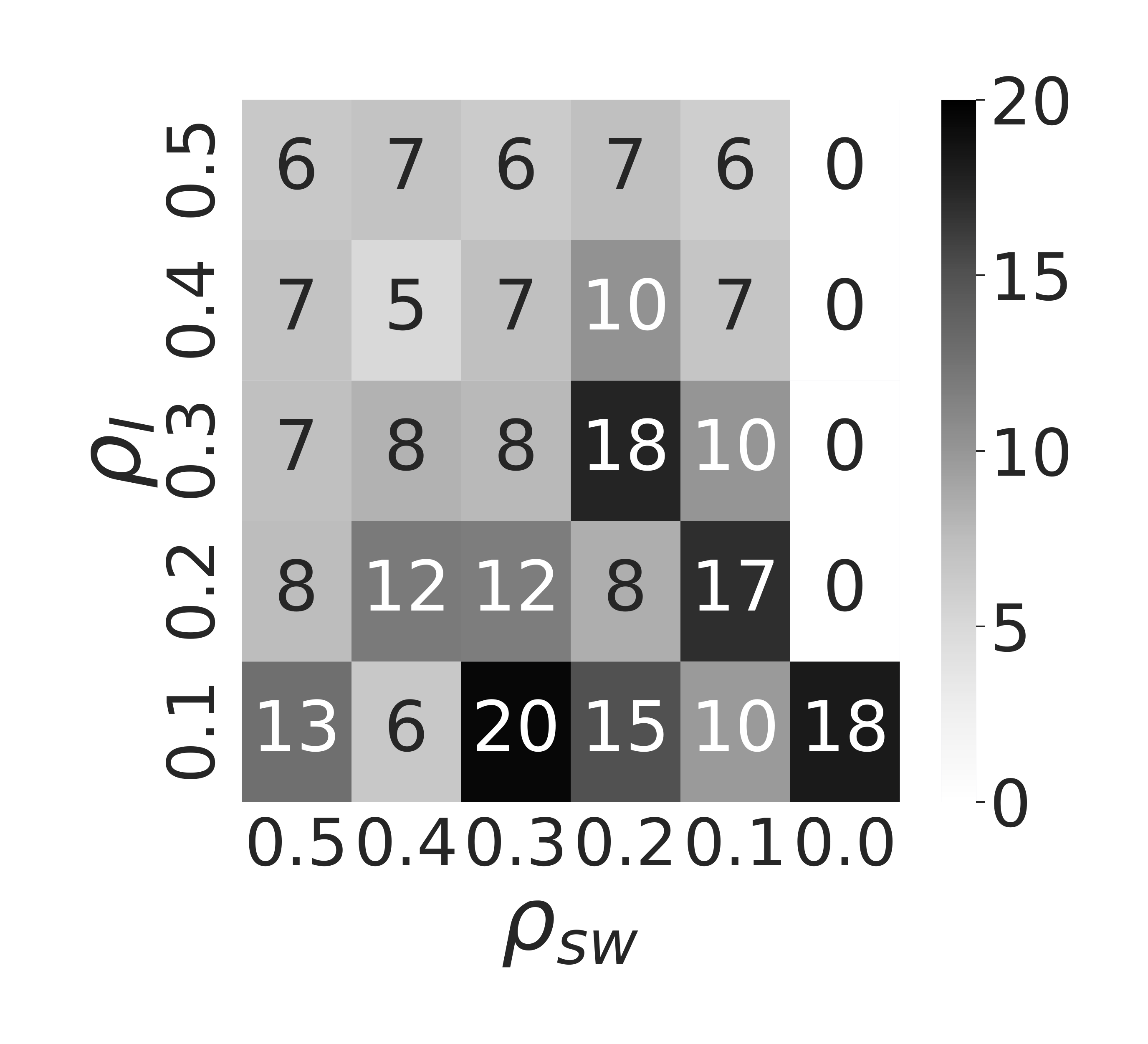}&
         \includegraphics[width=0.25\textwidth]{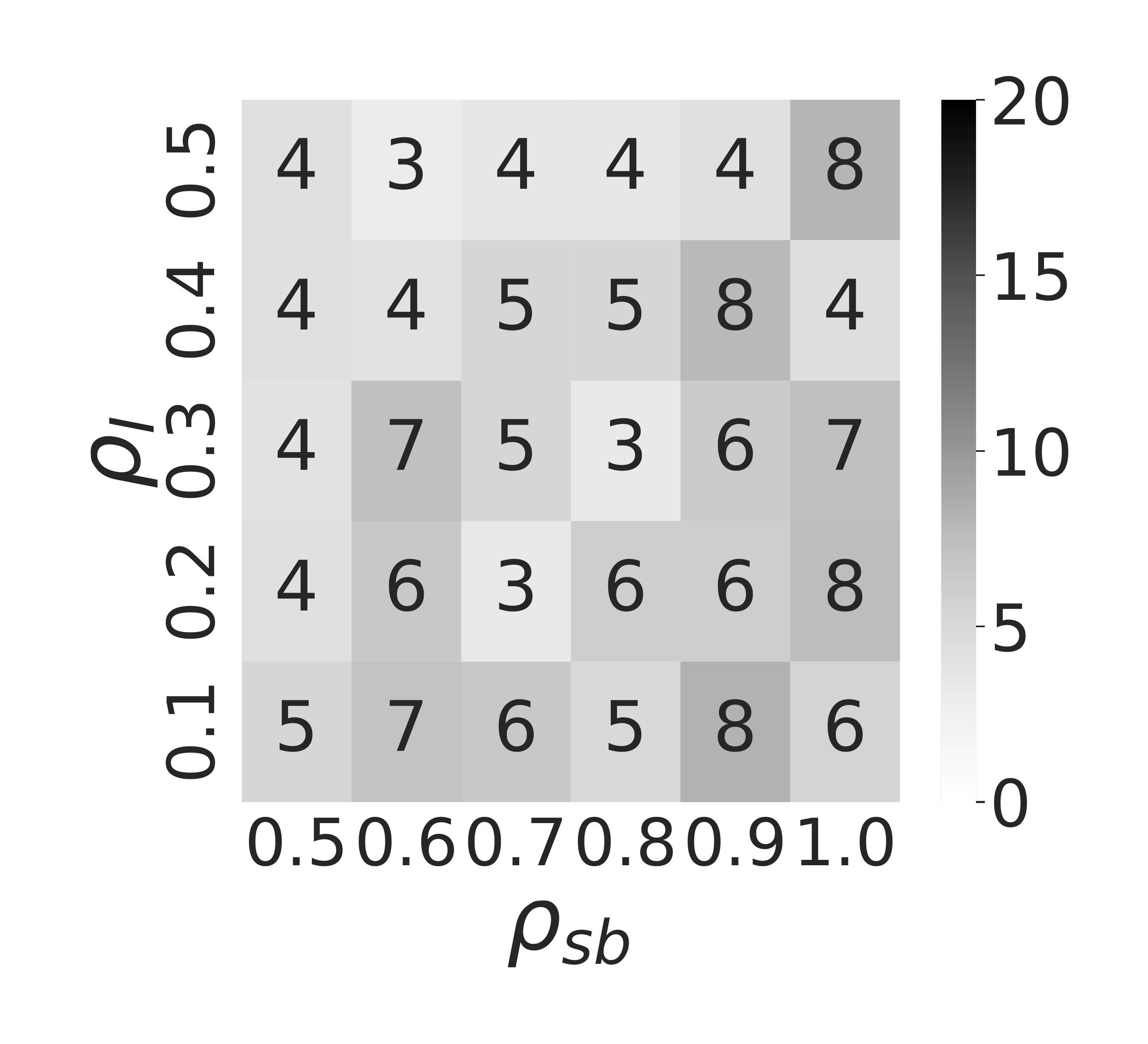}&
         \includegraphics[width=0.25\textwidth]{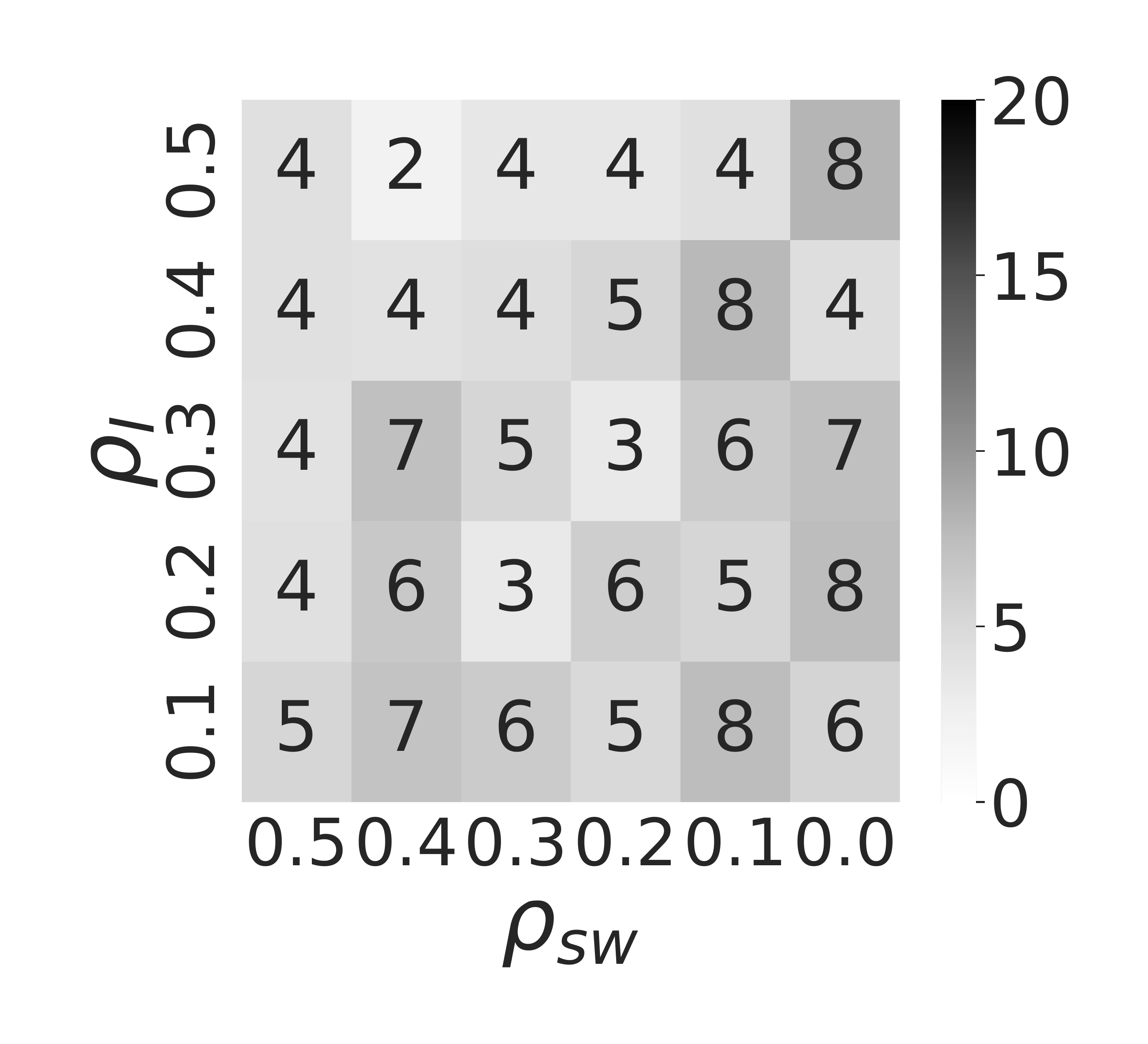}\\
         (a) & (b) & (c) & (d)\\
         \includegraphics[width=0.25\textwidth]{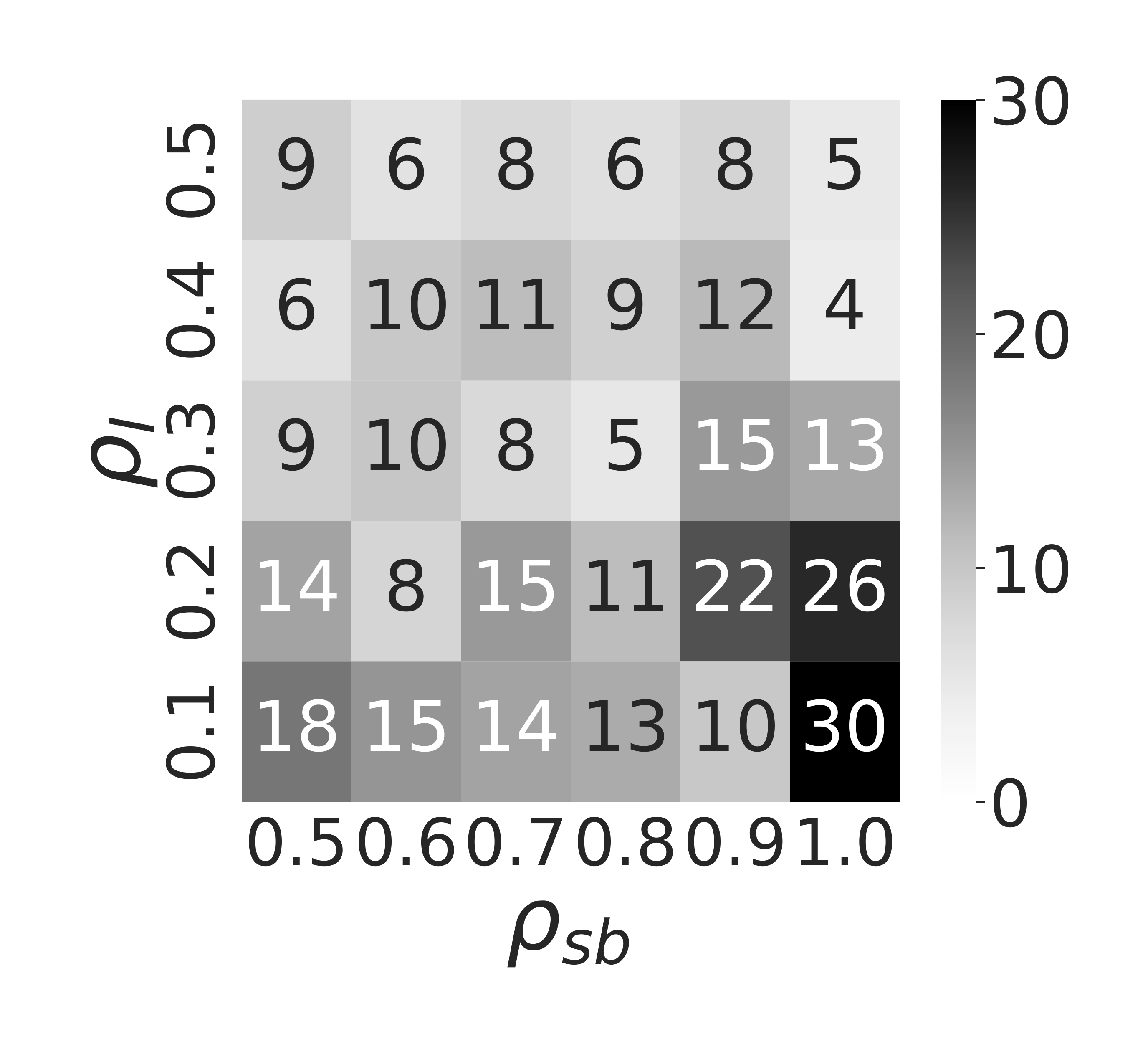}&
         \includegraphics[width=0.25\textwidth]{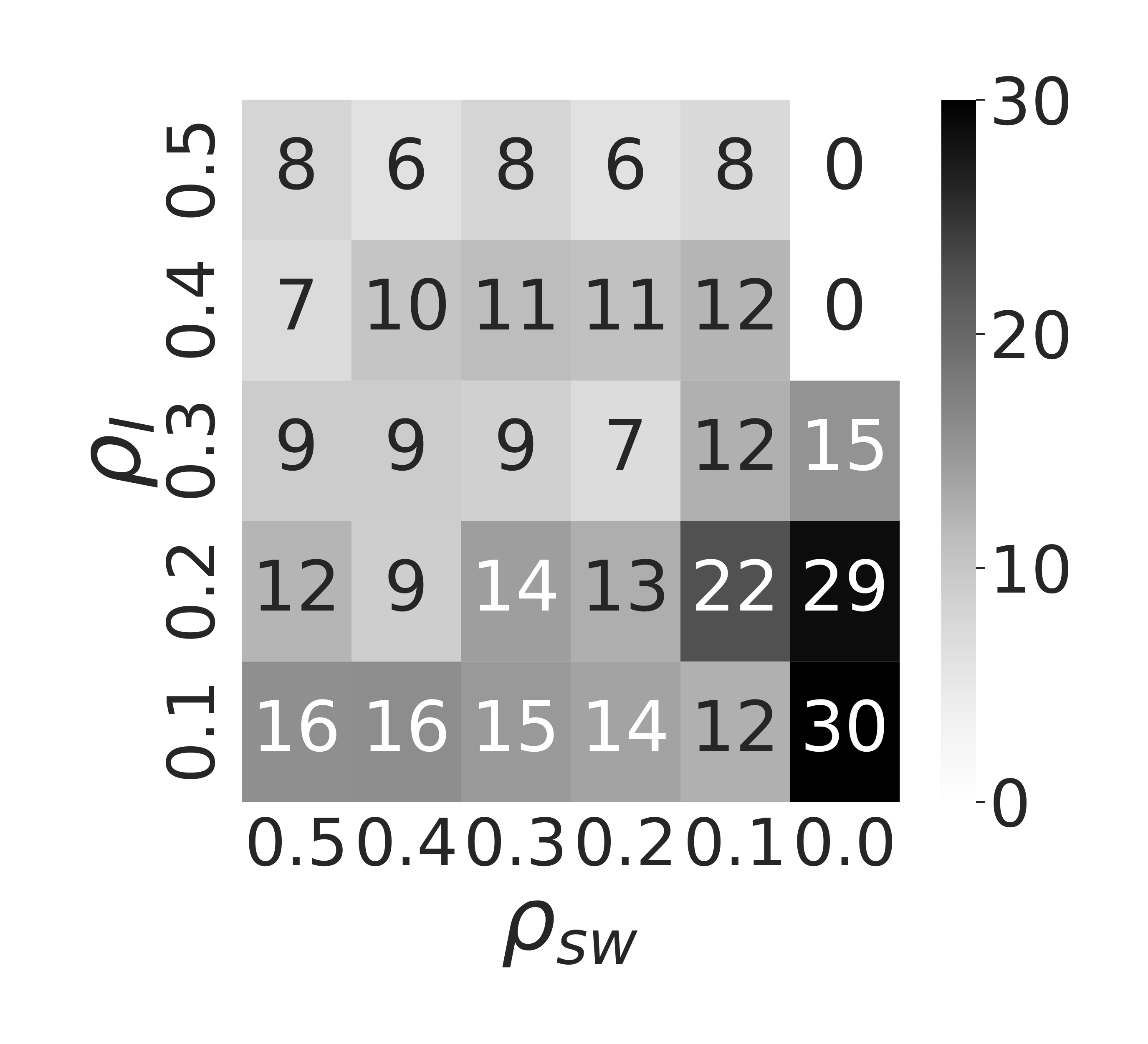}&
         \includegraphics[width=0.25\textwidth]{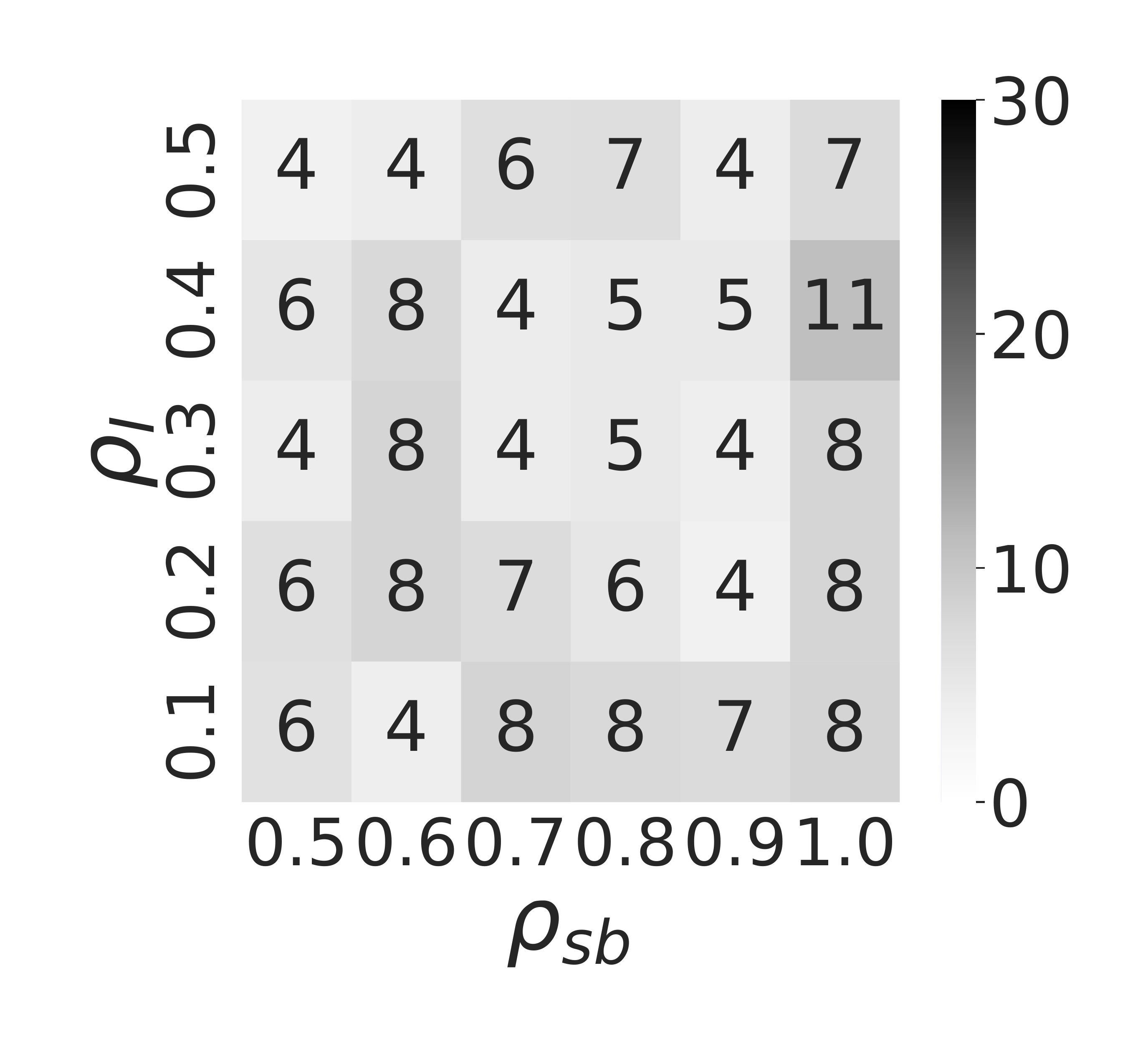}&
         \includegraphics[width=0.25\textwidth]{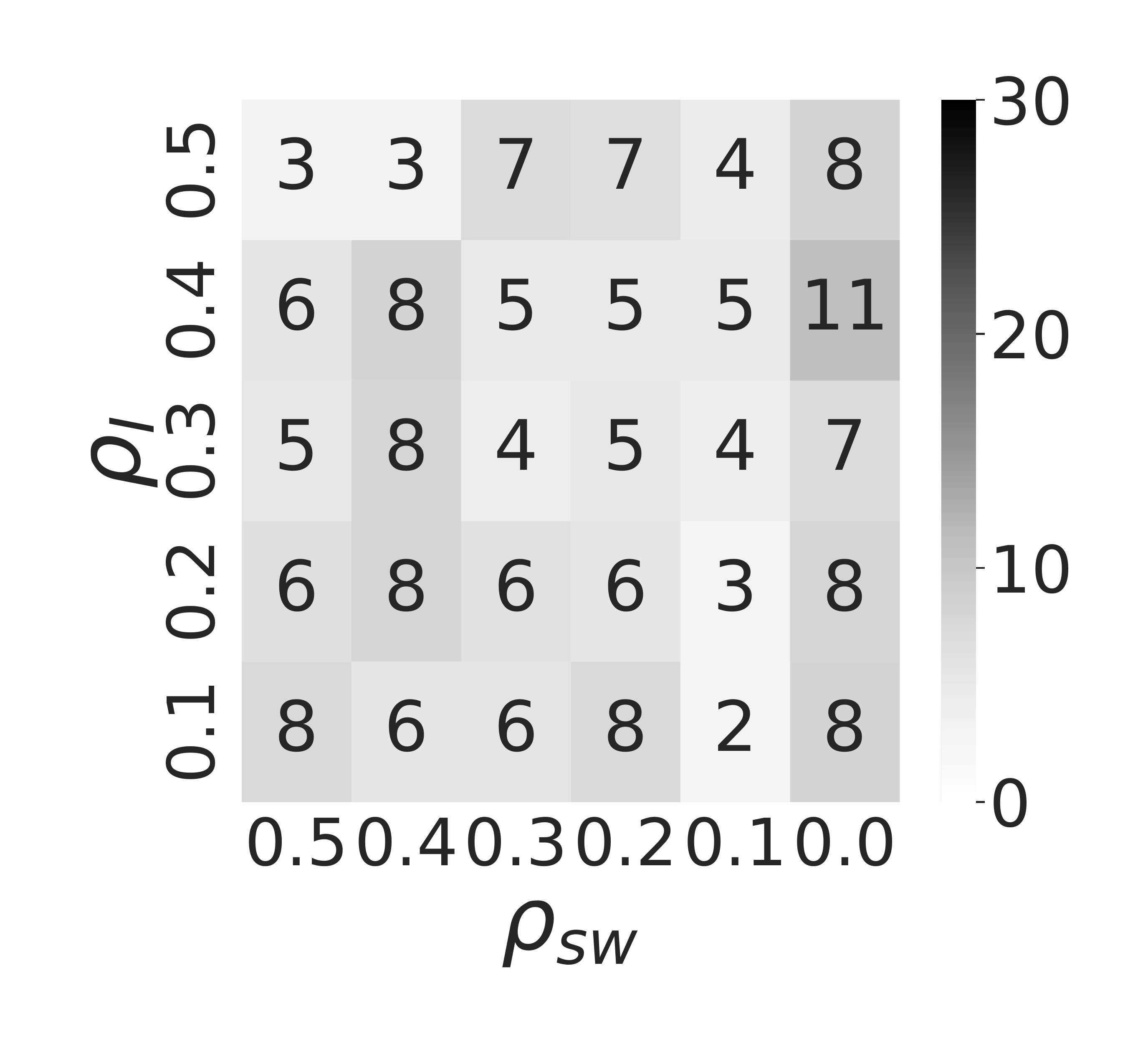}\\
         (e) & (f) & (g) & (h)
     \end{tabular}
     \caption{Graphs showing the interquartile range over 20 trials of the number of robots on each site at $t=$ 30,000 s (end of trial). See the caption of Fig.~\ref{fig:median} for more details.}
     \label{fig:variabilityBetweenRuns}
\end{figure}
Figure~\ref{fig:variabilityBetweenRuns} shows the interquartile ranges of the number of robots on each site at \unit[$t=30,000$]{s} for both approaches for a swarm size of $N=50$ (graphs in top row) and $N=100$ (graphs in bottom row). For low proportions of informed robots ($\rho_{I}< 0.3$), our approach shows a slightly higher variability in the final distribution of the swarms (Figs.~\ref{fig:variabilityBetweenRuns}a,~\ref{fig:variabilityBetweenRuns}b,~\ref{fig:variabilityBetweenRuns}e, and~\ref{fig:variabilityBetweenRuns}f) than the approach in~\cite{Firat2020Group-sizeRegulationInSelf-OrganisedAggregation} (see Figs.~\ref{fig:variabilityBetweenRuns}c,~\ref{fig:variabilityBetweenRuns}d,~\ref{fig:variabilityBetweenRuns}g, and~\ref{fig:variabilityBetweenRuns}h). However, at higher proportions of informed robots($\rho_{I} > 0.3$), the variability is roughly the same for the two approaches in the majority of the parameter configurations. 

\begin{figure}[h]
\centering
\begin{minipage}[c]{0.4\textwidth}
\includegraphics[width=\textwidth]{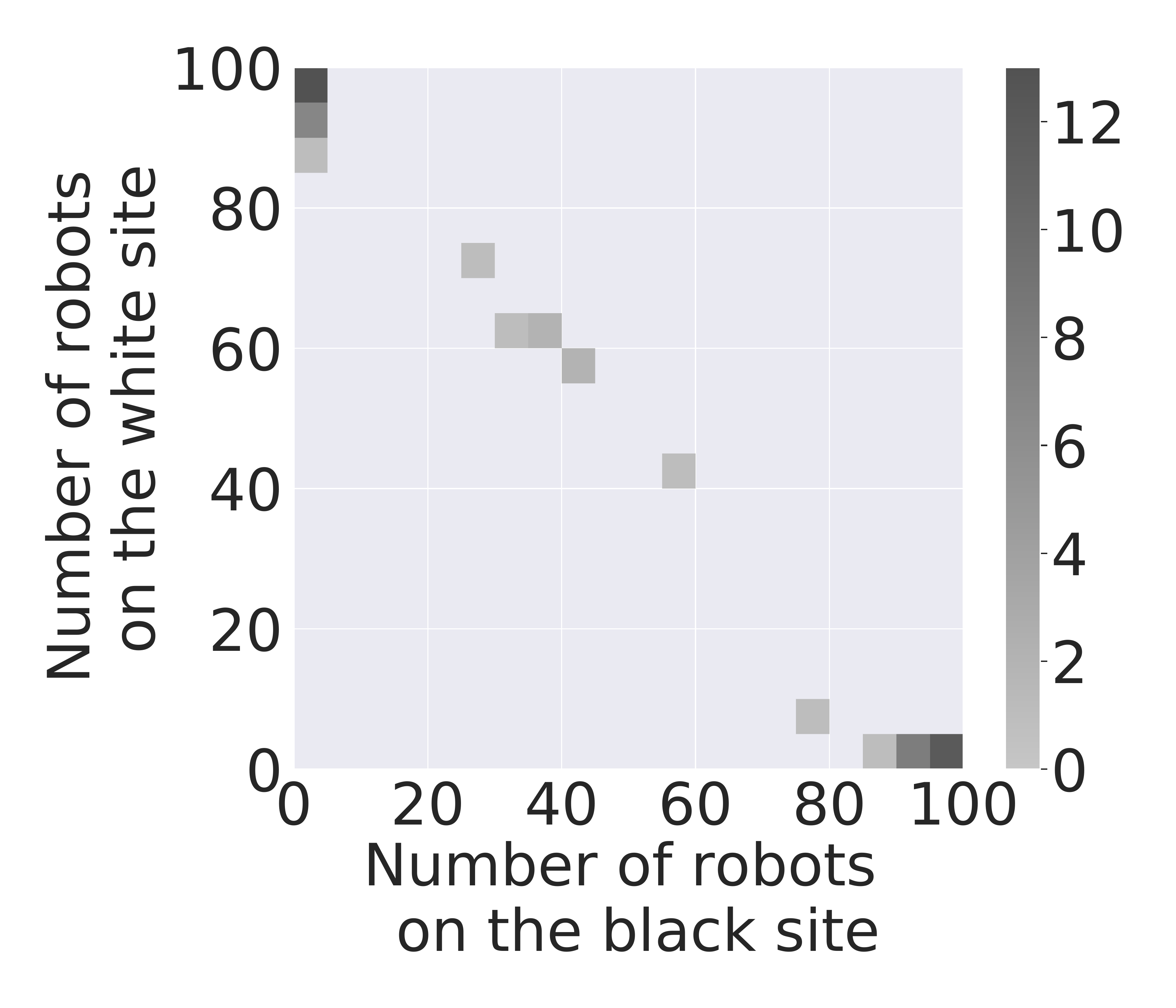}
\end{minipage}
\begin{minipage}[c]{0.4\textwidth}
\caption{Graphs showing the frequency distribution of the number of robots on each aggregation site, at the end of 50 runs, with each trial lasting $t=$ 30,000 s, and  without informed robots in the swarm ($\rho_{I} = 0$). The swarm size is $N=100$.}
\label{fig:symmetry}
\end{minipage}
\end{figure}

With our modified approach, we have run further tests with a swarm entirely made of non-informed robots (i.e., $\rho_{I}=0$). The removal of informed robots from the swarm prevents the model  illustrated in~\cite{Firat2020Group-sizeRegulationInSelf-OrganisedAggregation} from generating robots' aggregates, as the transition from state random walk $\mathcal{RW}$ to the state stay $\mathcal{S}$ (i.e., resting on the aggregation site) of non-informed robots is triggered by the perception of an aggregation site populated by informed robots. As shown in Fig.~\ref{fig:symmetry}, in our model, a swarm without informed robots can break the environmental symmetry by repeatedly forming a single aggregate on either the black or the white aggregation site. The median of the number of robots not resting on any site at the end of the experiment is 4 with an interquartile range of 5. This shows that the modifications we made to the system as originally illustrated  in~\cite{Firat2020Group-sizeRegulationInSelf-OrganisedAggregation} enlarges the swarm behavioral repertoire without loss of performance with respect to the results shown in~\cite{Firat2020Group-sizeRegulationInSelf-OrganisedAggregation}. 
The results of our tests show that our simplified model generates aggregates that match equally fine or even better than in~\cite{Firat2020Group-sizeRegulationInSelf-OrganisedAggregation} the expected distributions of robots on each aggregation site for each tested combination of values of $\rho_{I}$, $\rho_{sb}$, and $\rho_{sw}$. We show, however, that our approach generates a slightly higher between-trials variability than the one in~\cite{Firat2020Group-sizeRegulationInSelf-OrganisedAggregation}. This is due to the fact that when every robot can start the formation of clusters in the sites, as it is in our approach, clusters composed exclusively of non-informed robots can also appear. When the proportion of informed robots is low, the disruptive effect of these types of aggregates on the desired aggregation dynamics increases as well as with the between trials variability. We also show that, in the absence of informed robots in the swarm, the robots break the environmental symmetry by repeatedly generating a single aggregate on either the black or the white site. Since any aggregation behavior in the absence of informed robots is precluded to the swarm as modelled in~\cite{Firat2020Group-sizeRegulationInSelf-OrganisedAggregation}, we conclude that the modifications introduced by us improve the robustness and the behavioral flexibility of the swarm.

\section{Conclusions}
\label{sec:conclusion}
We proposed a simplified version of an existing aggregation method using informed individuals in a swarm of robots. We have presented the results of a comparative study that quantitatively evaluates the effectiveness of two different algorithms in driving the aggregation dynamics in swarms of heterogeneous robots made of informed and non-informed robots. The original approach, illustrated in~\cite{Firat2020Group-sizeRegulationInSelf-OrganisedAggregation}, is based on a finite-state machine controller by which individual robots only transit from state random walk $\mathcal{RW}$ (corresponding to random diffusion in the environment) to state Stay $\mathcal{S}$ (corresponding to resting on an aggregation site) when they end up on an aggregation site populated by informed robots. Moreover, the robots transit from state Stay $\mathcal{S}$ to state Leave $\mathcal{L}$ (corresponding to leaving the aggregation site) with a probability that depends on the variation in the perceived number of informed robots at the site during resting, thus requiring memory, however limited. The functional characteristics of this finite-state machine are supported by a communication protocol in which only informed robots can send signals and non-informed robots can receive signals. We have replicated the study illustrated in~\cite{Firat2020Group-sizeRegulationInSelf-OrganisedAggregation} with a simplified finite-state machine controller in which the robots rest on a site regardless of the presence of informed robots, and they leave the site with a memoryless probability that depends on the current perceived local density at the aggregation site of any type of robots. Moreover, we have re-established a functional equivalence in communication capabilities between informed and non-informed robots. That is, in our approach, there is no distinction in communication capabilities between the types of robots, both can send and receive messages that signal their presence at an aggregation site. This implementation choice allows the robots to estimate the robot density in their neighborhood without any need of distinguishing between robot types. Our comparative tests show that the swarms controlled by our approach generate aggregation dynamics that are never less performing than~\cite{Firat2020Group-sizeRegulationInSelf-OrganisedAggregation}, and in some experimental conditions they are closer to the target robot distributions than the one observed with the approach illustrated in~\cite{Firat2020Group-sizeRegulationInSelf-OrganisedAggregation}. We showed that, our approach, contrary to the one introduced in~\cite{Firat2020Group-sizeRegulationInSelf-OrganisedAggregation}, suffers from a slightly higher between-trials variability in the number of robots resting at each aggregation site at the end of a trial. We also show that our approach can generate a larger set of aggregation dynamics than in~\cite{Firat2020Group-sizeRegulationInSelf-OrganisedAggregation}, since in the absence of informed robots the swarm controlled with our simplified approach systematically forms a single aggregate in one of the aggregation sites, while the swarm controlled by the approach in~\cite{Firat2020Group-sizeRegulationInSelf-OrganisedAggregation} never aggregates. Based on the ``Occam's razor'' principle of parsimony, we claim that due to its simplicity, effectiveness, and robustness to a larger set of operating conditions, our approach should be favored over the one introduced in~\cite{Firat2020Group-sizeRegulationInSelf-OrganisedAggregation} to control the aggregation dynamics using informed robots.

Future research directions will consider setups with three or more aggregation sites to verify if, with our approach, the expected distributions of robots at the aggregation sites are attained in more complex environments. We also plan to validate the approach illustrated in this paper with the physical robot Kilobots~\cite{Rubenstein2014KilobotLowCostRobot} using the Kilogrid platform~\cite{Valentini2018KilogridANovelExperimentalEnvironment}. This will allow the study of the convergence time of the system and the possible speedup of the dynamics~\cite{Sion2022}.

\subsubsection*{Acknowledgements.}
This work was supported by Service Public de Wallonie Recherche under grant n° 2010235 - ARIAC by DIGITALWALLONIA4.AI; 
by the European Research Council (ERC) under the European Union’s Horizon 2020 research and innovation
programme (grant agreement No 681872); and by Belgium’s Wallonia-Brussels Federation through the ARC Advanced Project GbO (Guaranteed by Optimization). A.\,Reina and M.\,Birattari acknowledge the financial support from the Belgian F.R.S.-FNRS, of which they are Charg\'{e} de Recherches and Directeur de Recherches, respectively.
%
%
%
%
\bibliographystyle{splncs04}
\bibliography{mybibliography}
%
%




%
\end{document}